\documentclass[letterpaper]{article} 
\usepackage{aaai2026}  
\usepackage{times}  
\usepackage{helvet}  
\usepackage{courier}  
\usepackage[hyphens]{url}  
\usepackage{graphicx} 
\usepackage{natbib}  
\usepackage{caption} 
\usepackage{algorithm}
\usepackage{algorithmic}
\usepackage{booktabs}
\usepackage{amsmath}
\nocopyright
\usepackage{newfloat}
\usepackage{listings}
\DeclareCaptionStyle{ruled}{labelfont=normalfont,labelsep=colon,strut=off} 
\floatstyle{ruled}
\newfloat{listing}{tb}{lst}{}
\floatname{listing}{Listing}
\title{Beyond Scaling Law: A Data-Efficient Distillation Framework for Reasoning}
\author{
    Xiaojun Wu\equalcontrib\textsuperscript{\rm 1}, 
    Xiaoguang Jiang\equalcontrib\textsuperscript{\rm 1}, 
    Huiyang Li\textsuperscript{\rm 1}, 
    Jucai Zhai\textsuperscript{\rm 1}, 
    Dengfeng Liu\textsuperscript{\rm 1}, 
    Qiaobo Hao\textsuperscript{\rm 1}, \\
    Huang Liu\textsuperscript{\rm 1}, 
    Zhiguo Yang\textsuperscript{\rm 1}, 
    Ji Xie\textsuperscript{\rm 1}, 
    Ninglun Gu\textsuperscript{\rm 2},
    Jin Yang\textsuperscript{\rm 2},
    Kailai Zhang\textsuperscript{\rm 2},
    Yelun Bao\textsuperscript{\rm 2},
    Jun Wang\textsuperscript{\rm 2}
}
\affiliations{
    \textsuperscript{\rm 1}Zhongxing Telecom Equipment(ZTE), China \\
    \textsuperscript{\rm 2}China Mobile Communications Group Co Ltd\\

    AIM@zte.com.cn, \{guninglun, yangjin, zhangkailai, baoyelun, wangjunwl\}@chinamobile.com

}

\usepackage{bibentry}

\begin{document}

\maketitle

\begin{abstract}

Large language models (LLMs) demonstrate remarkable reasoning capabilities in tasks such as algorithmic coding and mathematical problem-solving. 
Recent methods have improved reasoning through expanded corpus and multistage training combining reinforcement learning and supervised fine-tuning. Although some methods suggest that small but targeted dataset can incentivize reasoning via only distillation, a reasoning scaling laws is still taking shape, increasing computational costs. To address this, we propose a data-efficient distillation framework (DED) that optimizes the Pareto frontier of reasoning distillation.
Inspired by the on-policy learning and diverse roll-out strategies of reinforcement learning, the key idea of our approach is threefold: (1) We identify that benchmark scores alone do not determine an effective teacher model. Through comprehensive comparisons of leading reasoning LLMs, we develop a method to select an optimal teacher model. (2) While scaling distillation can enhance reasoning, it often degrades out-of-domain performance. A carefully curated, smaller corpus achieves a balanced trade-off between in-domain and out-of-domain capabilities. (3) Diverse reasoning trajectories encourage the student model to develop robust reasoning skills.
We validate our method through evaluations on mathematical reasoning (AIME 2024/2025, MATH-500) and code generation (LiveCodeBench), achieving state-of-the-art results with only 0.8k carefully curated examples, bypassing the need for extensive scaling. Our systematic analysis demonstrates that DED outperforms existing methods by considering factors beyond superficial hardness, token length, or teacher model capability. This work offers a practical and efficient pathway to advanced reasoning while preserving general capabilities.
\end{abstract}


\section{Introduction}
\begin{figure}[t]
\centering
\includegraphics[width=0.45\textwidth]{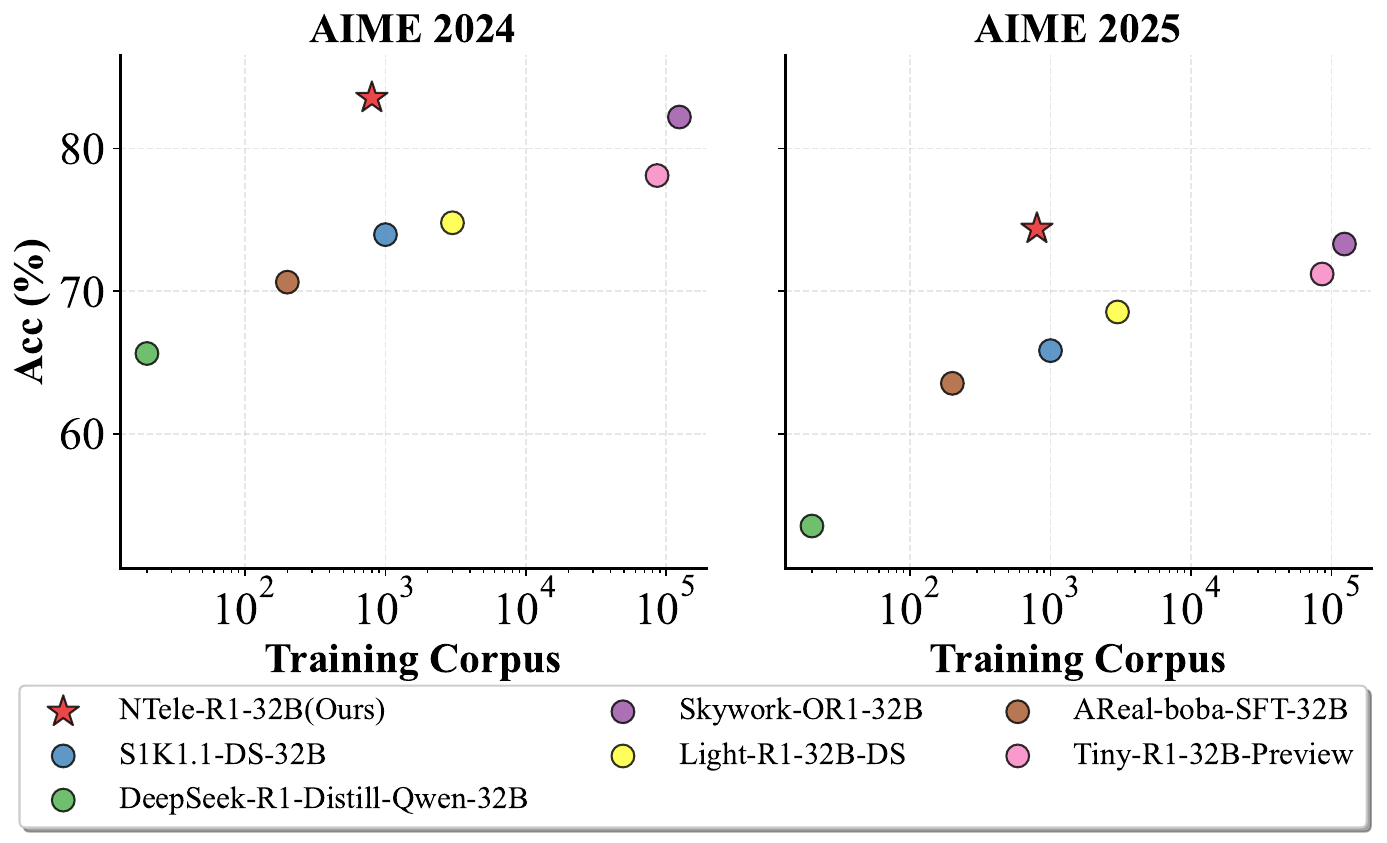} 
\caption{The performance on AIME 2024/2025 varies with the scale of the training corpus. Models fine-tuned from DeepSeek-R1-Distill-Qwen-32B exhibit a potential reasoning scaling law. Our model, NTele-R1-32B, breaks out of this trend and advances the Pareto frontier.}
\label{fig:pareto}
\end{figure}
In the last few months, large language models (LLMs) with Chain-of-Thought (CoT)\cite{wei_chain--thought_2023} reasoning emerge as the most potential pathway to achieve Artificial General Intelligence (AGI). 
In the pursuit of emergence of reasoning capabilities, two key techniques\cite{deepseek-ai_deepseek-r1_2025}, Reinforcement Learning with Verifiable Reward (RLVR)  and Supervised Fine-tuning (SFT) from distilled reasoning trajectories, attract the most attention of the community.
DeepSeek-R1-Zero\cite{deepseek-ai_deepseek-r1_2025} showed that large scale RLVR, together with the proposed Group Relative Policy Optimization (GRPO)\cite{shao_deepseekmath_2024}, could incentivize reasoning ability from scratch. Meanwhile, the open-sourced DeepSeek-R1-Distill-Qwen series models and relative studies\cite{novasky_team_sky-t1_2025,bespoke_labs_bespoke-stratos_2025,muennighoff_s1_2025,ye_limo_2025} showed that distillation might be a more practical way compared to RLVR when there exists a powerful reasoning LLM as the teacher. 
To take a step forward, recent distillation methods foster the reasoning of LLMs in two directions: (1) Enlarge the distilled CoT corpus, combined with one-stage selection. (2) Multistage training with iterative RLVR and distillation\cite{he_skywork_2025,wen_light-r1_2025}. A potential scaling law takes shape as illustrated in Figure~\ref{fig:pareto}. 

To lift up the scaling curve, researchers have proposed several metrics to filter high-quality examples from the distilled CoT corpus\cite{muennighoff_s1_2025}, including the difficulty, token length and diversity in domains.
These achievements raises a natural question: \textbf{Can we skip out of the potential scaling law?} Specifically, can we push the boundary of reasoning ability with limited examples? 

Inspired by the great improvement and systematic analysis of reinforcement learning, we found that three key factors remain underexplored in the area of CoT distillation: the strategy to select the teacher model, the impact on out-of-domain (OOD) LLM capabilities and the diversity of the question level.
To investigate, we conducted comprehensive experiments on the three aspects and constructed a data-efficient distillation (DED) framework. 

Our framework integrates three key innovations: (1) A practical strategy to select the teacher model;
(2) A practical compression of number of questions to reduce the damage to the OOD capabilities. 
(3) Diverse problem-solving trajectories for each question.
To demonstrate the effectiveness of our framework, we trained and open-sourced NTele--32B-V1, the state-of-the-art (SOTA) reasoning models of its parameters. we evaluate our model on the most commonly used reasoning benchmarks. As illustrated in Figure~\ref{fig:pareto}, our model greatly outperformed other baselines, especially considering the limited scale of training corpus.
In addition, we conducted comprehensive ablation studies and systematic analysis of our key innovations.

In summary, our contributions are:
\begin{itemize}
\item A novel DED framework that introduces two key factors that greatly impact the CoT distillation in the first time; 
\item An open-sourced SOTA reasoning model that greatly increase the reasoning scaling curve; 
\item Comprehensive ablation results and analysis demonstrating practical suggestions to CoT distillation.
\end{itemize}

\section{Related Work}









\textbf{Reasoning Corpus Synthesis. }
The CoT approach, which decomposes responses into finer-grained reasoning steps, has become the predominant technique for synthesizing high-quality training data in mathematics\cite{yu_metamath_2024,luo_wizardmath_2025} and coding\cite{luo_wizardcoder_2025} domains. To further enhance corpus quality, researchers have developed methods such as Tool-Integrated Reasoning\cite{gou_tora_2024, yang_qwen25-math_2024} and execution feedback\cite{xu_kodcode_2025,yang_evaluation_2024}, significantly improving the accuracy of the generated output. Following OpenAI's introduction of the o1 slow-thinking paradigm, research has begun synthesizing long-CoT corpora for imitation learning, successfully achieving Test-Time Scaling effects\cite{min_imitate_2024}. In particular, DeepSeek's breakthrough in enabling small models to achieve remarkable performance gains through large-scale knowledge distillation has attracted extensive research attention\cite{deepseek-ai_deepseek-r1_2025}, prompting a detailed exploration of distillation-based methodologies for synthesizing corpora of the R1 class and their underlying mechanisms\cite{elie_bakouch_open_2025,penedo_codeforces_2025,inclusionai_areal-boba-sft-32b_2025,he_skywork_2025}.

\noindent \textbf{Data-efficient SFT Practices. }
Research on LIMA\cite{zhou_lima_2023} demonstrates that the model can learn specific response formats with merely 1,000 carefully curated samples for SFT. This underscores the potential of SFT to fully utilize data and shows its ability to generalize to task domains absent from training data. Subsequently, the S1\cite{muennighoff_s1_2025} study extends the methodology to mathematical reasoning tasks by leveraging reasoning-capable LLMs such as DeepSeek-R1 as teacher models for distillation. This approach enables chat models to reliably switch into reasoning mode, achieving performance comparable to models fine-tuned with large-scale datasets. Further investigations by LIMO\cite{ye_limo_2025} and Light-R1\cite{wen_light-r1_2025} identified corpus difficulty as a crucial factor influencing SFT efficiency. These findings strongly indicate that carefully curated corpora enable highly efficient data utilization through SFT. Our research aims to further explore the limits of SFT utilization.

\noindent \textbf{Analysis of CoT distillation. }
Recent studies have demonstrated that the incorporation of detailed explanations can significantly enhance a model's ability to assimilate novel knowledge\cite{sun_how_2025}. Due to the multi-perspective analysis and step-by-step reasoning characteristics of slow thinking, SFT proves to be highly effective in enabling LLMs to master the reasoning patterns of slow thinking\cite{hu_why_2025,fu_srft_2025,liu_acereason-nemotron_2025}. On the other hand, the key to the slow thinking paradigm lies in logical keywords, which guide the model's reasoning direction and substantially influence the final accuracy. Fine-tuning can significantly reduce the information entropy of such critical tokens, increasing their frequency and allowing LLMs to easily adopt the slow thinking mode\cite{wang_beyond_2025,pang_token_2025,goncharov_complexity-aware_2025,cheng_reasoning_2025,lin_rho-1_2025}.

Notably, the primary implementation of SFT relies on distillation of the teacher model. However, the long distillation responses generated by different teacher models often exhibit two notable issues: (1) containing redundant reasoning steps, and (2) displaying significant stylistic discrepancies. These problems adversely affect the efficiency of knowledge transfer to the student model\cite{luo_deconstructing_2025}.


\section{Method}

\begin{figure*}[t]
\centering
\includegraphics[width=0.9\textwidth]{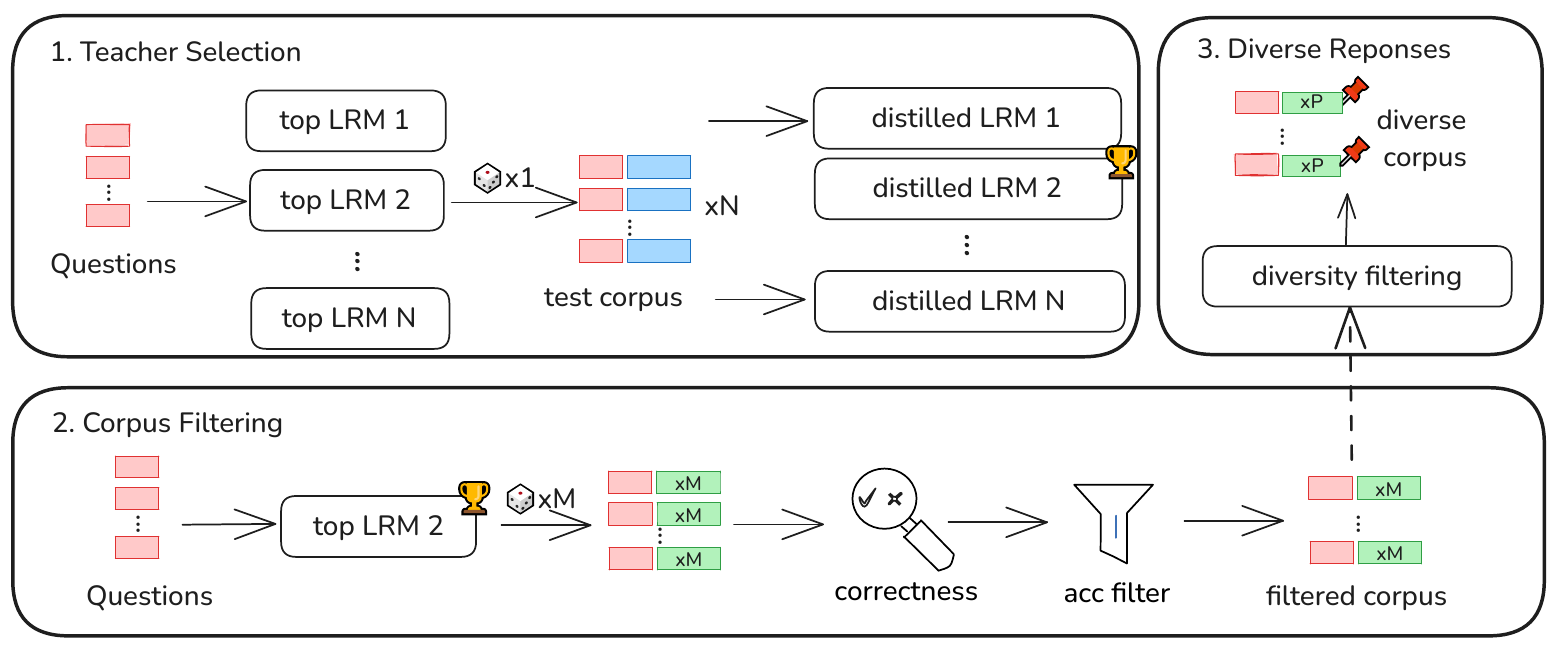} 
\caption{Overview of our data-efficient distillation framework.}
\label{fig:overall}
\end{figure*}

Our DED framework is designed to address the challenge of distillation with extremely limited number of examples. In other words, our framework dedicates to maximize the gains in reasoning capabilities, given the limited domain examples. To achieve this, our distillation framework, as illustrated in Figure~\ref{fig:overall} implements three sequential stages:



\textbf{Teacher Selection.} Given the limited questions, the first stage of our framework is to select a proper teacher model. Unlike previous methods, we do not take the most capable LLM in the reasoning benchmarks as the default. Instead, we argue that teaching ability does not always align with raw model strength. A smoke test is necessary to select the teacher. Specifically, we collect the best large reasoning models (LRMs) as candidates and sample one CoT response for each question. After the sampling, we conduct quick distillation tests on the student model and evaluate them on reasoning benchmarks. Then we choose the teacher according to the scores.

\textbf{Corpus Filtering.} Given the selected teacher, the second stage sampled the $M$ trajectories for each question. The sampled responses are used as follows. 
\begin{itemize}
\item \textbf{Quality and correctness check.} The sampled responses undergo strict quality checks, including filtering based on length (removing \( len > 16k \) samples), format (ensuring strict pairing of \( <think>...</think> \) delimiters), and correctness (verifying ground truth via rule verification and LLM-as-a-judge).
\item \textbf{Question Compression. }Hard samples from the verified corpus are identified on the basis of the pass rate of the student model. Specifically, the questions that are too easy for the student model are filtered out, whose pass rate is larger than a certain threshold.
\end{itemize}
\textbf{Diversity responses.} Inspired by the diverse roll-out strategy of reinforcement learning, we enhance the diversity of the trajectories for each question to achieve superior reasoning capabilities. 
Specifically, we calculate the Levenshtein distance among the trajectories and select the farthest $P$ responses for each question to ensure a diverse set.




\section{Experiments and Analysis}
\subsection{Setup}
The datasets, models, and evaluation settings related to our experiments are detailed below.

\textbf{Dataset. }Mathematical seed corpus we use is s1k\cite{muennighoff_s1_2025}, a data set of 1,000 examples of diverse, high-quality, and difficult problems. This dataset is distinguished by its exceptional quality and complexity, allowing robust evaluation and direct comparison with existing work. For the code seed, we select 1,000 programming challenges from CodeForce\cite{penedo_codeforces_2025} because of their complexity and community recognition.

\textbf{Model. }The DED framework adopts a teacher-student architecture.  We employ DeepSeek-R1\cite{deepseek-ai_deepseek-r1_2025} and QwQ-32B\cite{team_qwq-32b_2025} as teacher models, two popular reasoning models of different scales and architectures. For the student model, we deploy the widely adopted DeepSeek-R1-Distill-Qwen-32B (DS-32B), a mid-scale architecture balancing performance efficiency and computational economy.

\textbf{Benchmarks and Evaluation. }We evaluate the capabilities of the model in mathematical reasoning, programming, and general performance using the following benchmarks. 
\begin{itemize}
\item AIME 2024\cite{maa_aime_2024} : 30 integer answer problems from the 2024 American Invitational Mathematics Examination (AIME), testing advanced mathematical problem solving.
\item AIME 2025\cite{maa_aime_2024} : 30 novel problems from the 2025 AIME exams, evaluating forward-looking mathematical reasoning.
\item MATH500\cite{hendrycks_measuring_2021} : 500 multistep math word problems that span algebra, geometry, calculus, and number theory, requiring symbolic manipulation.
\item GPQA Diamond\cite{rein_gpqa_2023} : Features diverse, high-difficulty tasks in mathematics, reasoning, and commonsense knowledge to assess model generalization and robustness
\item LiveCodeBench (LCB)\cite{jain_livecodebench_2024} : A dynamically updated, contamination-free coding benchmark sourcing tasks from LeetCode, Codeforces, and AtCoder (Feb. 2025-May 2025 snapshot).
\end{itemize}

To ensure result stability, we compute the pass@1 metric over 16 evaluation runs for all benchmarks except GPQA Diamond and MATH500.

\textbf{Training Details. }
We build a training code based on the Llama Factory framework\cite{zheng_llamafactory_2024} with a context window of 16,384 tokens, batch size of 48, and learning rate of \(1\times10^{-5}\) using AdamW optimization. We used one \(8 \times H800\) node to train the 32B model for 10 epochs, and it took an estimated 9 hours for 1,000 pieces of data.

\begin{table*}[!ht]
    \centering
    \scalebox{0.85}{
        \begin{tabular}{lcccccccccccc}
        \toprule
        \textbf{} & \multicolumn{3}{c}{\textbf{AIME 2024}} & \multicolumn{3}{c}{\textbf{AIME 2025}} &\multicolumn{3}{c}{\textbf{Math500}} &\multicolumn{3}{c}{\textbf{GPQA Diamond}}  \\
        \cmidrule(lr){2-4} \cmidrule(lr){5-7} \cmidrule(lr){8-10} \cmidrule(lr){11-13}
        \textbf{Model} &
        \textbf{ACC} & \textbf{Length} & \textbf{ $\Delta$Acc} & \textbf{ACC} & \textbf{Length} & \textbf{ $\Delta$Acc} & \textbf{ACC} & \textbf{Length} & \textbf{ $\Delta$Acc} & 
        \textbf{ACC} & \textbf{Length} & \textbf{ $\Delta$Acc} \\
        \midrule
        DS-32B & 65.63  & 10171  & - & 53.54 & 12514 & - & 89.8 & 2360 & - & 62.10 & 5094 & - \\
        * DeepSeek-R1 & 73.96 & 11255 & +8.33 & 65.83 & 12930 & +12.29 & 94.6 & 3576 & +4.8 & 64.65 & 7506 & +2.55 \\
        * QwQ-32B & \textbf{79.58} & 14096 & +\textbf{13.95} & \textbf{73.22} & 16275 & +\textbf{19.68} & \textbf{95.6} & 4910 & +\textbf{5.8} & 67.17 & 9726 & +5.07 \\
        * Qwen3-32B & 75.00 & 14675 & +9.37 & 69.73 & 16371 & +16.19 & 94.2 & 5242 & +4.4 & 62.63 & 9682 & +0.53 \\
        * Qwen3-235B-A22B & 78.75 & 13991 & +13.12 & 68.75 & 16277 & +15.21 & 90.8 & 7201 & +1.0 & \textbf{67.68} & 11071 & +\textbf{5.58} \\
        
        \cmidrule(lr){1-13}
        Qwen2.5-32B-Instruct & 17.50  & 10171  & - & 10 & 12514 & - & 81.0 & 2360 & - & 48.50 & 5094 & - \\
        * DeepSeek-R1 & 57.08 & 14698 & +39.58 & 46.88 & 12930 & +36.88 & \textbf{93.2} & 3744 & +\textbf{12.2} & 62.63 & 8601 & +14.13 \\
        * QwQ-32B & \textbf{67.50} & 16335 & +\textbf{50.00} & \textbf{61.04} & 17633 & +\textbf{51.04} & 92.8 & 4912 & +11.8 & \textbf{63.64} & 9659 & +\textbf{15.14} \\
        
        \cmidrule(lr){1-13}
        phi-4 & 17.92  & 1539  & - & 17.29 & 1201 & - & 79.6 & 652 & - & 51.01 & 627 & - \\
        * DeepSeek-R1 & 42.92 & 12345 & +25.00 & 32.91 & 11224 & +15.62 & 88.4 & 3825 & +8.8 & \textbf{62.50} & 8241 & +\textbf{11.49} \\
        * QwQ-32B & \textbf{50.62} & 12433 & +\textbf{32.70} & \textbf{41.67} & 12827 & +\textbf{24.38} & \textbf{90.0} & 4910 & +\textbf{10.4} & 58.59 & 9209 & +7.58 \\  
        \bottomrule
        \end{tabular}
    }
    \caption{Comparisons of different teacher-student pairs. The symbol \textbf{*} means the model is distilled from the corresponding teacher. \textbf{Bold} values highlight the highest scores. Results show that QwQ-32B serves as a more effective teacher than DeepSeek-R1, Qwen3-32B, and Qwen3-235B-A22B, even though these models achieve stronger performance on math benchmarks. This suggests that the most capable LRM in downstream tasks is not necessarily the optimal teacher.}
    \label{t:teacher}
\end{table*}

\subsection{Main Results}

We compare the models trained on the basis on the DED framework with other representative works to verify efficiency and performance. 

\begin{itemize}
\item DeepSeek-R1, QwQ-32B, Qwen3-32B and Qwen3-235B-A22B are our teacher models and serve as our performance references.
\item Light-R1-32B-DS\cite{wen_light-r1_2025} is similar to our work. It is synthesized on the basis of a large amount of open source data and selects the difficult 3,000 data for SFT training.
\item AReal-boba-SFT-32B\cite{inclusionai_areal-boba-sft-32b_2025} is also a DS-32B trained that uses only 200 data samples and a small amount of computing resources.
\item Skywork-OR1-32B\cite{he_skywork_2025} employs the GRPO\cite{shao_deepseekmath_2024} algorithm to train on a dataset that includes more than 120,000 problems.
\end{itemize}

\begin{table}[!ht]
    \centering
    \scalebox{0.8}{
        \begin{tabular}{lcccc}
        \toprule
        \textbf{Model} & \textbf{AIME 2024} & \textbf{AIME 2025} & \textbf{Math500}\\ 
        \midrule
        DeepSeek-R1 & 79.2 & 70 & 97.3 \\ 
        Qwen3-235B-A22B & 85.7 & 81.5 & 98.0 \\
        \midrule
        QwQ-32B & 76.25 & 67.30 & 94.6\\ 
        Qwen3-32B & 78.75 & \underline{73.33} & \underline{94.6}\\ 
        Light-R1-32B-DS & 74.79 & 68.54 & 92.0\\ 
        AReal-boba-SFT-32B & 70.63 & 63.54 & 88.8\\ 
        Skywork-OR1-32B & \textbf{82.2} & 73.3 & -\\ 
        NTele-R1-32B-Math & \underline{81.87} & \textbf{77.29} & \textbf{95.2}\\ 
        \bottomrule
        \end{tabular}
    }
    \caption{Comparisions among different methods on math benchmarks}
    \label{t-table1}
\end{table}
Table~\ref{t-table1} shows the evaluation results of models trained with different methods on AIME 2024, AIME 2025 and MATH500. 
In mathematical reasoning tasks, the DED framework shows remarkable effectiveness using a small number of samples. Using a diversity-enhanced corpus synthesized from merely 200+ queries (representing only 20\% of the raw sample pool), DED achieves notable accuracy rates of \textbf{81.87\%} and \textbf{77.29\%} on the AIME 2024 and AIME 2025 benchmarks, respectively improving by 16.24\% and 23.75\% over the base. Notably, this performance not only surpasses those models that use the same base but a larger synthetic corpus, but also surpasses the QwQ-32B and DeepSeek-R1 teacher models. These results robustly validate the efficacy and data efficiency of our proposed approach in low-resource mathematical reasoning scenarios.



\textbf{Teacher Specialization. }To investigate the impact of teacher models, we perform extensive experimental validation. Using DS-32B as the student model, we additionally select two models from the Qwen3 series, Qwen3-32B and Qwen3-235B-A22B, as teacher models and evaluate the performance of the trained models on various tasks.

Table\ref{t:teacher} shows that student models trained with the distillation corpora of the QwQ-32B teacher model consistently outperform those trained with Deepseek-R1 on all evaluation benchmarks, while also exceeding the performance of models distilled from Qwen3-32B and Qwen3-235B-A22B. Additionally, we observe that models distilled from Qwen-series teacher models generally produce longer responses compared to models distilled from Deepseek-R1. The impact of corpus and response length on performance is analyzed in section~\ref{subsec:ana}.

To explore potential differences among student models, we conduct comparative experiments using Qwen2.5-32B-Instruct, Phi-4. The results consistently show that the student models distilled from QwQ-32B as the teacher model are stronger than the student models distilled from DeepSeek-R1 as the teacher model on the benchmark test. Notably, after training on 1,000 corpus samples, Qwen2.5-32B-Instruct and Phi-4 exhibited substantial performance improvements on AIME 2024 and AIME 2025. These findings challenge conventional scaling laws, providing strong evidence for the efficacy of our proposed distillation approach.

\begin{table}[!ht]
    \centering
    \scalebox{0.75}{
        \begin{tabular}{lccc|c}
        \toprule
        \textbf{Model} & \textbf{AIME 2024} & \textbf{AIME 2025} & \textbf{Math500} & \textbf{Num}\\ 
        \midrule
        DS-32B & 65.63 & 53.54 & 89.8 & -\\ 
        R1-Math-DS-32B & 73.96 & 65.83 & 94.6 &1000 \\ 
        QwQ-Math-DS-32B & 79.58 & 73.22 & \underline{95.6} & 1000 \\ 
        + Right & \underline{80.00} & \underline{75.21} & 95.0 & 830 \\ 
        + Right-Hard & 79.37 & 72.29  & \textbf{95.6} & 237 \\ 
        + Right-Hard-Diverse & \textbf{81.87} & \textbf{77.29} & 95.2 & 965 \\ 
        \bottomrule
        \end{tabular}
    }
    \caption{Results on models distilled with corpus from different stages. The questions are from s1 dataset. The last column shows the number of training samples}
    \label{t-corpus}
\end{table}

\textbf{Compression and Diversity. }Experimental results from Table~\ref{t-corpus} demonstrate that models trained in quality-filtered corpora exhibit enhanced performance compared to those trained on unfiltered corpora, underscoring the influence of corpus quality on training results. Due to the exceptional capability of the teacher model, only 170 corpus samples were filtered, indicating that the unfiltered corpus already possessed high quality. Consequently, the observed performance improvement remains modest.

To further optimize the training corpus, we compressed it to approximately 200 samples by selecting hard examples, reducing the corpus size by 75\%. This compression led to a slight decline in the model performance. We then expanded the corpus by applying diversity enhancement techniques to these selected difficult examples and found that models trained on the expanded corpus outperformed models trained on the full dataset. These findings highlight the effectiveness of hard-example compression, demonstrating that the quality of training data is more critical than its quantity.
Furthermore, diversity enhancement successfully surpasses the performance of models trained on the full corpus, challenging the traditional reliance on large-scale datasets. This demonstrates the effectiveness of quality-driven efficient corpus construction methods and provides valuable insights for designing compact and efficient training datasets in resource-constrained settings.

\subsection{DED for Code Generation}
To evaluate the generality and effectiveness of the DED framework for reasoning tasks in code generation, we used a corpus of 1,000 code samples and fully reused the DED process for training and evaluation.

\begin{table}[!ht]
    \centering
    \scalebox{0.75}{
        \begin{tabular}{lccc|c}
        \toprule
        \textbf{Model} & \textbf{LCB easy} & \textbf{LCB medium} & \textbf{LCB hard} & \textbf{Num}\\
        \midrule
        DS-32B & 88.37 & 54.93 & 10.39 & \\
        R1-Code-DS-32B & 80.37 & 57.57 & 15.62 & 1000\\
        QwQ-Code-DS-32B & 85.39 & 66.35 & 25.27 &1000 \\
        +Right & \underline{89.31} & \underline{66.58} & \underline{26.72} & 846\\
        +Right-Hard & 83.72 & 61.17 & 23.13  &230 \\
        +Right-Hard-Diverse & \textbf{90.26} & \textbf{70.19} & \textbf{28.43} & 925 \\
        \bottomrule
        \end{tabular} 
    }
    \caption{Results on models distilled with corpus from different stages. The questions are sampled from CodeForce dataset. The last column shows the number of training samples}
    \label{t-code}
\end{table}

To facilitate the evaluation of performance variations at different difficulty levels, we categorized the LCB evaluation set into easy, medium, and hard subsets. As shown in the evaluation results in Table~\ref{t-code}, the DED framework has been successfully adapted to the domain of code generation. Given the high accuracy of the base model on easy questions and the focus of training on challenging problems, the primary improvements are observed in the medium and hard subsets. Compared to Deepseek-R1, QwQ-32B demonstrates superior performance as a teacher model for code-related tasks. Using 230 samples augmented to 925 via diversity enhancement, our DED framework achieves SOTA performance on the LCB benchmark, outperforming the teacher model, demonstrating its efficiency and versatility in few-shot learning across domains.


We also performed experiments to evaluate the impact of training with a mixed corpus of code and math on task performance and cross-domain generalization. To maintain consistency with the size of the individual corpora, we selected a mixed corpus of 400+ math samples and 400+ code-related samples (for a total of 800+ samples).

Results of Table~\ref{t-mixed} reveals that training exclusively on either mathematical or code-related corpora yields improvements in the other domain. When combining half of each corpus for mixed training, our approach achieves SOTA performance in both code generation and mathematical tasks. These results highlight the superior efficacy of our method in enhancing cross-domain capabilities.

\begin{table*}[!ht]
    \centering
    \scalebox{1.0}{
        \begin{tabular}{lccccccc}
        \toprule
        \textbf{} & \multicolumn{4}{c}{\textbf{Math}} & \multicolumn{3}{c}{\textbf{Code}} \\
        \cmidrule(lr){2-5} \cmidrule(lr){6-8}
        \textbf{Model} & \textbf{AIME 2024} & \textbf{AIME 2025} & \textbf{Math500} & \textbf{GPQA Diamond} & \textbf{LCB easy} & \textbf{LCB medium} & \textbf{LCB hard} \\
        \midrule
        DeepSeek-R1 & 79.2 & 70 & \textbf{96.4} & \textbf{75.7} & - & - & - \\
        QwQ-32B & 76.25 & 67.30 & 94.6 & 63.60 & 92.44 & 68.63 & 23.05 \\
        DS-32B & 65.63 & 53.54 & 89.8 & 62.10 & 88.37 & 54.93 & 10.39 \\
        DS-32B-Math & 81.87 & \textbf{77.29} & \underline{95.2} & \underline{68.69} & \textbf{92.88} & 62.94 & 18.52 \\
        DS-32B-Code & 73.96 & 64.17 & 94.8 & 65.15 & 90.26 & \underline{70.19} & \underline{28.43} \\
        NTele-R1-32B & \textbf{83.54} & \underline{74.37} & \underline{95.2} & 67.68 & \underline{92.58} & \textbf{71.03} & \textbf{30.94} \\
        \bottomrule
        \end{tabular}
    }
    \caption{Performance of different models on math and code evaluation set}
    \label{t-mixed}
\end{table*}

We evaluated the impact of combined mathematical and code training on out-of-domain performance through a comprehensive assessment of general knowledge, common sense reasoning, and mathematical and coding benchmarks. The evaluation covers various benchmarks, including MMLU\cite{hendrycks_measuring_2021-1}, CMMLU\cite{li_cmmlu_2024}, C-EVAL\cite{huang_c-eval_2023}, BBH\cite{shi_language_2022}, MBPP, GSM8K\cite{cobbe_training_2021}, MATH\cite{hendrycks_measuring_2021}, and Aider’s Polyglot Benchmark.

\begin{table*}[!ht]
    \centering
    \scalebox{1}{
        \begin{tabular}{lcccccccc}
        \toprule
        \textbf{Model} & \textbf{MMLU} & \textbf{CMMLU} & \textbf{C-EVAL}  & \textbf{BBH} & \textbf{MBPP} & \textbf{GSM8K} & \textbf{MATH} & \textbf{Aider(paas@2)} \\ 
        \midrule
        DS-32B & 87.79 & 84.88 & 87.98 & 72.74 &89.60 & 92.57 & 88.28  & 12.4 \\ 
        NTele-R1-32B & \textbf{89.51} & \textbf{86.48}	&\textbf{88.82} & \textbf{74.99}	& \textbf{90.80}	& \textbf{95.98}	& \textbf{94.20} & \textbf{25.8} \\ 
        \bottomrule
        \end{tabular}
    }
    \caption{Model performance across diverse benchmarks after mathematical and code distillation}
    \label{t-moreeval}
\end{table*}

Table~\ref{t-moreeval} reveals that models trained on mixed mathematical and code corpora using the DED framework exhibit improved performance across all evaluation sets compared to the base. Notably, on the Aider benchmark, the score increased from 12.4 to 25.8, doubling the performance. These results strongly demonstrate the effectiveness and generalization of our method.

\subsection{More Analyses}
\label{subsec:ana}
In this section, we analyze the deep-seated reasons for the effectiveness of the DED framework from multiple perspectives, to provide reference for the widespread application of efficient extraction.



\textbf{Effect of Length. }As shown in Table~\ref{t:teacher}, models trained on the QwQ-32B sampled corpus outperform models trained on DeepSeek-R1 on all evaluation sets. To investigate the potential influence of training corpus characteristics, we calculate the token length differences between the two corpora. The results reveal that the QwQ-32B corpus has a higher average token length (12,431) compared to DeepSeek-R1 (9,877).

To further explore whether the longer training corpus contributes to improved performance through extended chains of thought in responses, we conducted an experiment by sampling multiple responses from QwQ-32B for the same query. Specifically, we compared the performance of models trained on the longest responses (average term length: 13,869) with the shortest responses (average term length: 8,202) to assess the impact of response length on performance.

\begin{table}[!ht]
    \centering
    \scalebox{0.8}{
        \begin{tabular}{lccc}
        \toprule
        \textbf{Model} & \textbf{Corpus} & \textbf{AIME 2024 / Len} & \textbf{AIME 2025 / Len}  \\ 
        \midrule
        R1-DS-32B & 9,877 & 73.96 / 11,255 & 65.83 / 12,930  \\ 
        QwQ-DS-32B & 12,431 & 79.58 / 14,219 & 73.22 / 16,275 \\ 
        QwQ-DS-32B-L & 13,869 & 78.12 / 14,767 & 74.42 / 16,449 \\
        QwQ-DS-32B-S & 8,202 & 77.71 /  12,696 & 74.58 / 14,732 \\ 
        \bottomrule
        \end{tabular}
    }
    \caption{Model performance distilled from corpora of different lengths}
    \label{t-length}
\end{table}



As presented in Table~\ref{t-length}, the results reveal that three corpora sampled from QwQ-32B display substantial variations in their length distributions. Notably, differences in response lengths are also observed on the evaluation set, yet the accuracy across these corpora remains remarkably consistent. Furthermore, a comparison between the training corpus lengths and evaluation performance of QwQ-DS-32B-S and R1-DS-32B demonstrates that QwQ-DS-32B-S, despite being trained on a corpus with a shorter average length, generates longer responses while achieving superior accuracy. 

Consequently, these findings suggest that neither the length of the corpus nor the response length serves as a dominant factor influencing distillation performance. This insight underscores the shortcomings of prior studies that rely solely on length as an indicator of difficulty or corpus quality, providing valuable guidance for corpus selection and quality evaluation.


\textbf{Token Entropy Analysis. }Token entropy analysis measures the uncertainty in the probability distribution of tokens generated by a language model at each step of its output.We analyze the differences between the corpora of two teacher models using token entropy. By comparing the differences in the overall token entropy across the corpora, we can gain insights into the affinity of student models for the distillation corpora derived from different teacher models.

\begin{figure*}[t]
\centering
\includegraphics[width=0.98\textwidth]{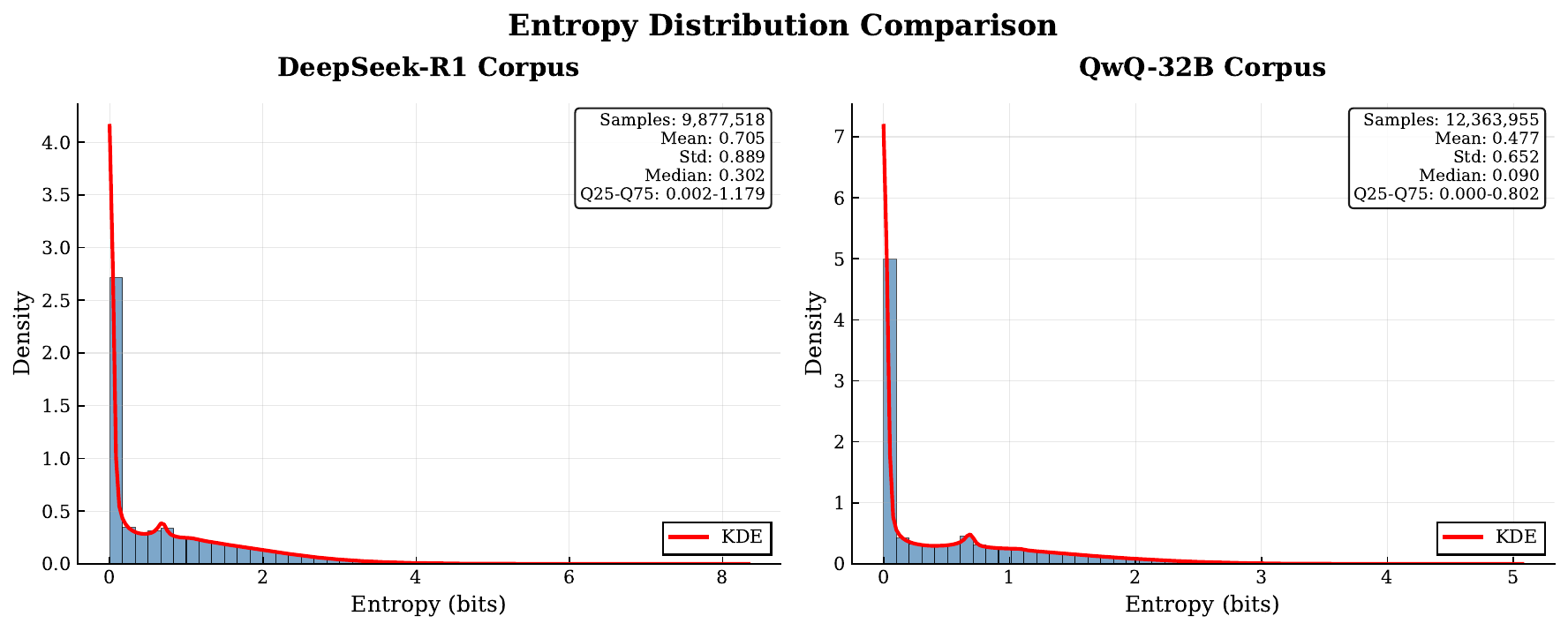} 
\caption{Comparison of token entropy distribution of teacher models. }
\label{fig:entropy}
\end{figure*}

As shown in Figure~\ref{fig:entropy}, The QwQ-32B corpus (mean: 0.477, median: 0.090) exhibits significantly lower token entropy compared to Deepseek-R1 (mean: 0.705, median: 0.302), reflecting a more predictable and structured token distribution. This reduced entropy has profound implications for both training dynamics and evaluation performance. The lower token entropy of the QwQ-32B corpus enhances the convergence of the student model by reducing the variability of the token sequence, promoting consistent learning of patterns relevant to the task. This structured corpus improves model robustness and generalization to OOD tasks, highlighting QwQ-32B’s efficacy as a teacher model for efficient knowledge distillation and strong performance across diverse benchmarks.


\textbf{PCA Shift Analysis. }
Recent studies\cite{xu_unlearning_2025,zheng_spurious_2025} show that PCA shift analysis serves as a robust and interpretable metric to evaluate changes in latent representations associated with improvements in task performance.
To further analyze the performance gap between different teacher model distillations, we used PCA shift analysis to examine the changes in model content features under different training corpora. Using the open-source method from the work\cite{huan_does_2025}, we computed the latent representation shifts for models trained with various teacher-distilled corpora.

\begin{figure*}[!ht]
\centering
\includegraphics[width=0.98\textwidth]{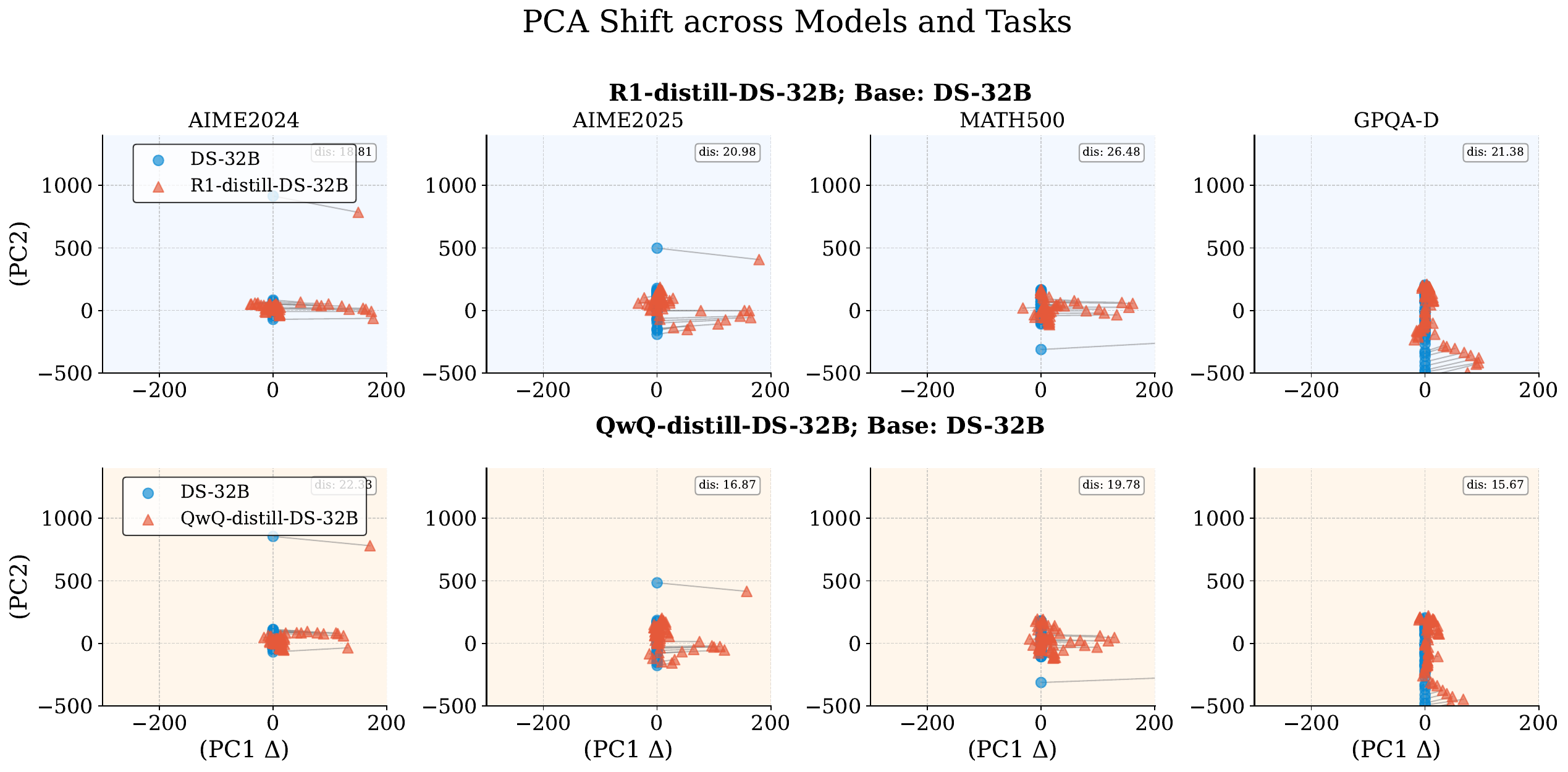} 
\caption{PCA offset of DS-32B across various teacher models and tasks. \(dis\) represents the Euclidean distance between the centroids of latent representations before and after training. Models trained on the QwQ-32B corpus exhibit smaller PCA offsets than DeepSeek-R1 across most tasks, indicating greater stability in their latent representations.}
\label{fig:pca}
\end{figure*}


The results in Figure~\ref{fig:pca} demonstrate that model distilled from QwQ-32B exhibit lower PCA shift on multiple evaluation benchmarks, including AIME2025, MATH500, and GPQA Diamond, compared to the model distilled from Deepseek-R1. This observation yields two critical insights. First, it confirms that the characteristics of the distilled corpora from different teacher models are significantly different, particularly in token probability distributions and token entropy, as previously noted. These corpus-specific variations likely contribute to the observed differences in representational stability, as captured by the PCA offset metric. Second, the reduced PCA offsets in models distilled from QwQ-32B indicate heightened stability in their latent representations, which is associated with superior generalization performance in OOD tasks. This finding highlights the key role of corpus affinity (i.e., the consistency between characteristics of the teacher model and the target task domain) in optimizing distillation results. Therefore, these results advocate careful consideration of corpus selection in knowledge distillation to maximize model robustness and generalization across different evaluation scenarios.

\section{Conclusion}
In this study, we challenge the dominant scaling-centric paradigm in distillation by introducing DED, a data-efficient framework designed to enhance reasoning capabilities. Through teacher model selection, curated corpus design, and diverse reasoning trajectories, DED achieves SOTA performance on mathematical reasoning (AIME 2024/2025, MATH-500) and code generation (LCB) benchmarks using only 800 curated examples. Our approach surpasses existing methods by pointing out that superficial factors such as teacher ability and token length are not the primary factors affecting distillation. Instead, we should focus on the token entropy of the corpus and the shift in latent representations space. This can enhance reasoning capabilities while achieving higher efficiency and OOD generalization. Future work will investigate DED's performance and generalization across a broader range of models and domains. Additionally, we aim to enhance the framework's interpretability to better understand its reasoning processes.

\bibliography{eval,model,general,distill,corpus}

\end{document}